\documentclass[10pt,twocolumn,letterpaper]{article}
\usepackage[pagenumbers]{cvpr} % To force page numbers, e.g. for an arXiv version
\definecolor{cvprblue}{rgb}{0.21,0.49,0.74}
\usepackage[pagebackref,breaklinks,colorlinks,allcolors=cvprblue]{hyperref}
\usepackage{color, soul}
\usepackage{booktabs}%表格宏包
\usepackage{makecell}
\usepackage{verbatim}
\usepackage{tabularray}

\usepackage{multirow}
\usepackage{xcolor}
\usepackage{ifthen}
\usepackage{microtype}
\usepackage{amsmath}
\usepackage{amssymb}
\usepackage{graphicx}
\usepackage{amsfonts}
\usepackage{array}
\usepackage{stfloats}
\usepackage{url}
\usepackage{ragged2e}
\usepackage{ulem}
 
\hypersetup{colorlinks}
\usepackage{caption}
\usepackage{subcaption}

\usepackage{enumitem}
\usepackage{engord}
\usepackage{fmtcount}
\usepackage{microtype}
\usepackage{balance}
\usepackage[super]{nth}
\renewcommand{\thefootnote}{}

\newcommand{\sysname}{\textit{HeadRouter}\xspace}

\definecolor{FanColor}{rgb}{0.8,0,0.8}

\definecolor{YuxinColor}{rgb}{0.54,0.18,0.88}

\newcommand{\nothing}[1]{}

\definecolor{figred}{rgb}{1,0,0}
\definecolor{figgreen}{rgb}{0,0.6,0}
\definecolor{figblue}{rgb}{0,0,1}
\definecolor{figpink}{rgb}{1,0.63,0.63}

\renewcommand{\paragraph}[1]{\textbf{#1}}

\def\paperID{8403} % *** Enter the Paper ID here
\def\confName{CVPR}
\def\confYear{2025}
\title{HeadRouter: A Training-free Image Editing \\ Framework for MM-DiTs by Adaptively Routing Attention Heads}

\author{
Yu Xu\textsuperscript{1,2},
Fan Tang\textsuperscript{1,2},
Juan Cao\textsuperscript{1,2}, 
Yuxin Zhang\textsuperscript{3,2}, 
Xiaoyu Kong\textsuperscript{4},
Jintao Li\textsuperscript{1}, \\
Oliver Deussen\textsuperscript{5}, 
Tong-Yee Lee\textsuperscript{6}
}

\begin{document}
\twocolumn[{%
\renewcommand\twocolumn[1][]{#1}%
\maketitle
\begin{center}
\url{https://yuci-gpt.github.io/headrouter/}
\end{center}
\begin{center}
    %\centering
    \captionsetup{type=figure}
    \includegraphics[width=\linewidth]{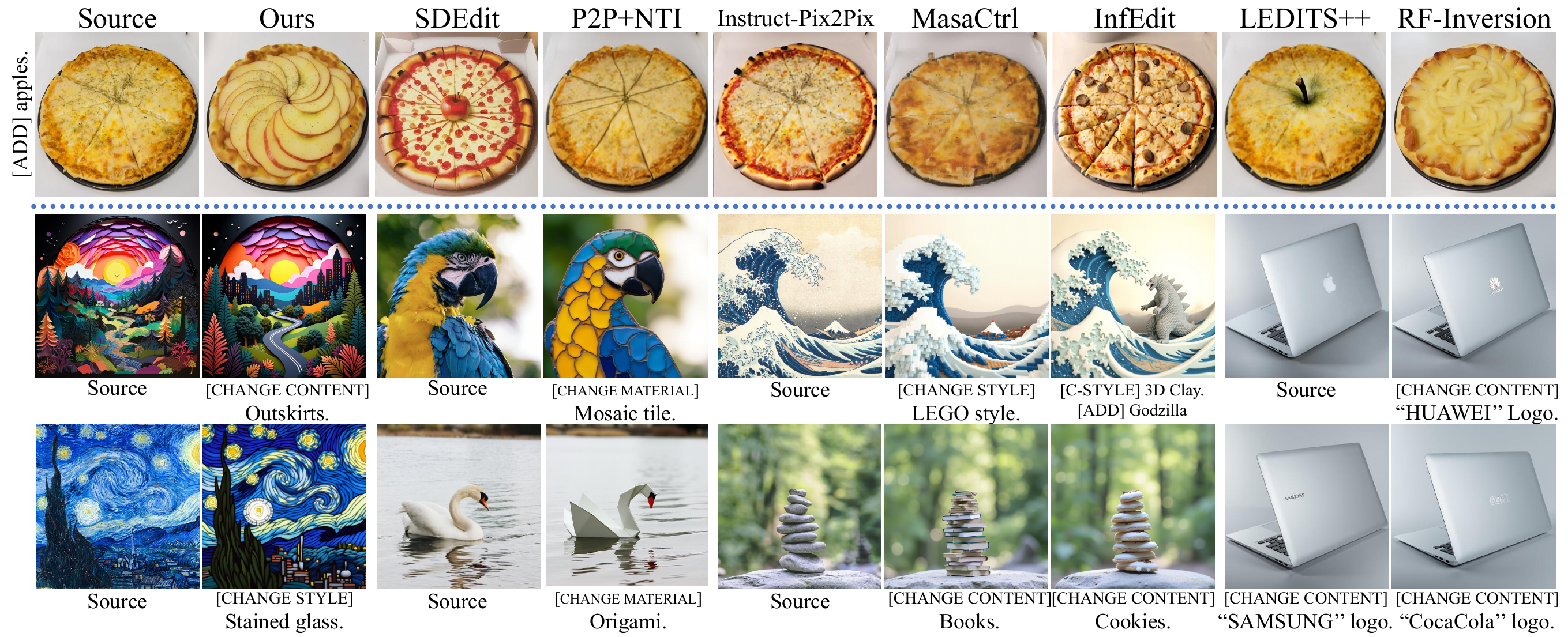}
    \vspace{-6mm}
    \caption{{Results of \textbf{HeadRouter} demonstrate accurate text-guided semantic representation while preserving consistency with the source image across diverse editing tasks.}}
   \label{fig:teaser}
\end{center}
}]

\renewcommand{\thefootnote}{}
\footnotetext{{
$^{1}$Institute of Computing Technology, Chinese Academy of Sciences; $^{2}$University of Chinese Academy of Sciences; $^{3}$Institute of Automation, Chinese Academy of Sciences; $^{4}$Beihang University; $^{5}$University of Konstanz; $^{6}$National Cheng-Kung University
}}

\begin{abstract}
Diffusion Transformers (DiTs) have exhibited robust capabilities in image generation tasks. 
However, accurate text-guided image editing for multimodal DiTs (MM-DiTs) still poses a significant challenge.
Unlike UNet-based structures that could utilize self/cross-attention maps for semantic editing, MM-DiTs inherently lack support for explicit and consistent incorporated text guidance, resulting in semantic misalignment between the edited results and texts. 
In this study, we disclose the sensitivity of different attention heads to different image semantics within MM-DiTs and introduce \sysname, a training-free image editing framework that edits the source image by adaptively routing the text guidance to different attention heads in MM-DiTs.
Furthermore, we present a dual-token refinement module to refine text/image token representations for precise semantic guidance and accurate region expression. 
Experimental results on multiple benchmarks demonstrate HeadRouter's performance in terms of editing fidelity and image quality.
\end{abstract}

\section{Introduction}
\label{sec:intro}

The introduction of the Diffusion Transformers~\cite{peebles2023scalable} (DiTs) architecture, which combines diffusion processes with transformers~\cite{dosovitskiy2020image}, has substantially augmented the scalability and performance of visual generation models (e.g., Sora~\cite{openai2024sora}, PixArt-$\alpha$~\cite{chen2024pixartalpha}, etc). 
Building upon these advances, the multimodal Diffusion Transformers (MM-DiTs) has been proposed and developed by SD3~\cite{esser2024scaling} and Flux~\cite{flux}. 
MM-DiTs employ a joint self-attention architecture that integrates image and text inputs, which effectively captures inter-feature relationships and generates a more aligned feature space, demonstrating outstanding performance in various downstream tasks such as controllable generation and personalized generation with enhanced scalability, generation quality, and accuracy.

Despite these advancements, the MM-DiTs architecture poses unique challenges to semantic image editing tasks. 
Traditional UNet~\cite{ronneberger2015u}-based diffusion models~\cite{rombach2022high,saharia2022photorealistic,podellsdxl} employ cross-attention mechanisms to incorporate text guidance, thereby generating text-image cross-attention maps.
These attention maps afford semantic control over the generated features, facilitating previous image editing methods to achieve various editing operations. 
Differently, MM-DiTs utilize joint self-attention, integrating text and image embeddings as input. 
This integration entangles multimodal features, eliminating explicit interactions between text and images, which presents challenges in effectively leveraging MM-DiTs for image editing tasks.

Several approaches endeavor to accomplish image editing based on DiTs/MM-DiTs.
Lazy DiT~\cite{nitzan2024lazy} requires training of both encoders and decoders, which renders it resource-intensive. 
RF-Inversion~\cite{rout2024semantic} puts forward an edit-friendly inversion method that leverages dynamic optimal control through a linear quadratic regulator. 
Nevertheless, accurately representing the semantic information required for text-guided editing remains a challenge within the context of RF-inversion.
Consequently, there is still a need for an accurate, high-quality image editing method that is applicable to MM-DiTs.

In this study, we delve into the inner structure of MM-DiTs to uncover the semantic representation properties in multi-head attention mechanisms.
By generating images from diverse text pairs with semantic differences and quantifying the similarity between different semantics and the output features of different heads, we showcase correlations between semantics and specific heads.
In contrast to the previous study~\cite{gandelsmaninterpreting}, which disclosed that certain attention heads capture specific image properties in the CLIP-ViT~\cite{dosovitskiy2020image}, we highlight that the various image semantics are adaptively distributed across different heads for MM-DiTs.
Meanwhile, we explore the feature interactions between text and image tokens within joint self-attention. 
Different from UNet, joint self-attention lacks cross-attention for introducing explicit text guidance. 
By analyzing the interaction mechanism between text and image tokens, we identify and extract critical regions in the joint self-attention map where text tokens influence image tokens. 

In light of these observations, we present \sysname, a training-free image editing framework for MM-DiTs. An instance-adaptive attention head router (\textbf{IARouter}) is put forward, which adaptively activates attention heads based on their semantic sensitivity, thereby enabling a more accurate representation of the edited specific images.
We further propose a dual-token refinement module (\textbf{DTR}) that employs self-enhancement of image tokens and rectified text token methods to enhance the representation of text-guided editing features in key regions and deep joint self-attention blocks. 
Besides, our method maintains time efficiency by avoiding additional modules or complex attention computations.
Our contributions can be summarized as follows:
\begin{itemize}
\item We provide insightful analysis into the influence of different attention heads on various editing semantics, as well as the interactions between text and image tokens in the cross-attention-free MM-DiTs.

\item We propose \sysname, a novel image editing method suitable for MM-DiTs, which includes an instance-adaptive router to enhance key attention heads for semantic representation and a dual-token refinement module for accurate text guidance and key region expression.

\item Experimental evaluation on multiple text-guided image editing benchmarks demonstrates that our approach yields more accurate regional, semantic, and attribute-wise editing effects across diverse tasks, surpassing state-of-the-art baseline methods.

\end{itemize}

\section{Related Work}
\label{sec:relate}

%-------------------------------------------------------------------------
\subsection{Diffusion transformers}
% Diffusion models have emerged as a powerful approach for image generation, with models like DDPM~\cite{ho2020denoising} demonstrating high-quality outputs. 
Integrating transformers into diffusion models leverages their ability to model long-range dependencies. 
Diffusion Transformers (DiTs)~\cite{peebles2023scalable} pioneered this integration for class-conditioned image generation, combining the strengths of diffusion processes and transformer architectures.
Building on DiT, models such as PixArt-$\alpha$~\cite{chen2024pixartalpha}, SD3~\cite{esser2024scaling}, and Flux~\cite{flux} extended this framework to text-conditioned image generation. 
% These models incorporate textual information, enabling more expressive and controllable synthesis. 
Specifically, SD3 and Flux utilize Multimodal Diffusion Transformers (MM-DiTs), entangling text and image modalities during training and inference. 
This entanglement enhances the interaction between textual and visual features, leading to images that better reflect the input text.
Despite these advancements, existing methods treat attention heads uniformly without exploiting their potential for semantic-specific representation. 
Our approach addresses this gap by assigning different attention heads to specific editing semantics in the diffusion transformer. 
By enhancing token representations with text prompts, we achieve more precise control over image semantics, resulting in images that more accurately align with the textual descriptions.

\subsection{Text-guided training-free image editing}
Training-free image editing approaches are considered fast and cost-effective since they eliminate the need for training or finetuning on data throughout the editing process~\cite{huang2024diffusion}.
% Previous work primarily employed two approaches: improving inversion and sampling methods for real image editing and utilizing attention mechanisms within UNet for editing. 
DDIM inversion~\cite{songdenoising} is a foundational approach for inversion-based methods~\cite{mokady2023null,huberman2024edit,garibi2024renoise} that leverages deterministic denoising steps to invert a real image back into noise. 
% Null-text inversion~\cite{mokady2023null} explores ``null'' or empty conditioning in diffusion models to enable precise reconstruction of real images for editing. 
% Methods such as DDPM Inversion~\cite{huberman2024edit} and LEDITS++~\cite{brack2024ledits++} incorporate information from the inversion stage to inform the sampling stage for reconstruction. 
% Renoise~\cite{garibi2024renoise} employs an iterative renoising mechanism to refine the approximation of a predicted point for enhanced image editing.
InfEdit~\cite{xu2024inversion} implies a virtual inversion strategy without explicit inversion in sampling.
However, MM-DiTs are mainly based on rectified flow models with ordinary differential equation~\cite{liuflow, albergobuilding, lipmanflow}, which is different from SDE-based diffusion models; thus, using the above method in MM-DiTs for image editing is less effective.  
RF-inversion~\cite{rout2024semantic} proposes a dynamic optimal control-based approach for inverting rectified flow models. 
However, the lack of exploration into the inherent mechanisms and characteristics of editing in MM-DiTs limits its ability.

Besides, attention modification techniques offer a direct and commonly used approach for training-free image editing~\cite{parmar2023zero,cao2023masactrl,tumanyan2023plug,liu2024towards}.
Prompt-to-prompt~\cite{hertzprompt} enables prompt-based image editing by aligning spatial relationships in cross-attention layers, ensuring consistency between the edited result and the source image. 
Guide-and-Rescale~\cite{titov2024guide} also uses a modified diffusion sampling process with self-guidance from attention maps.
CDS~\cite{nam2024contrastive} extracts the intermediate features of the self-attention layers and calculate loss to regulate structural consistency.
However, removing the cross-attention mechanism in MM-DiTs and joint input of text and image embeddings leads to entangled features of image and text, making it hard to explicitly utilize the cross-attention map for text-guided image editing. 
Therefore, we analyze the influence of text tokens on image tokens in MM-DiTs, identifying and extracting critical regions in the joint attention map to guide image editing.

% Pix2Pix-Zero~\cite{parmar2023zero} leverages cross-attention guidance to discover editing directions in the text embedding space.
% MasaCtrl~\cite{cao2023masactrl}, PnP~\cite{tumanyan2023plug} and FPE~\cite{liu2024towards} focus on attention feature replacement for consistency, with MasaCtrl and FPE modifying the key and value in self-attention and PnP manipulating spatial features and self-attention to inject features. 
\section{Analysis on Multiple Head Attention}
\label{sec:analysis}
\begin{figure}[!t]
  \centering
  \includegraphics[width=1\linewidth]{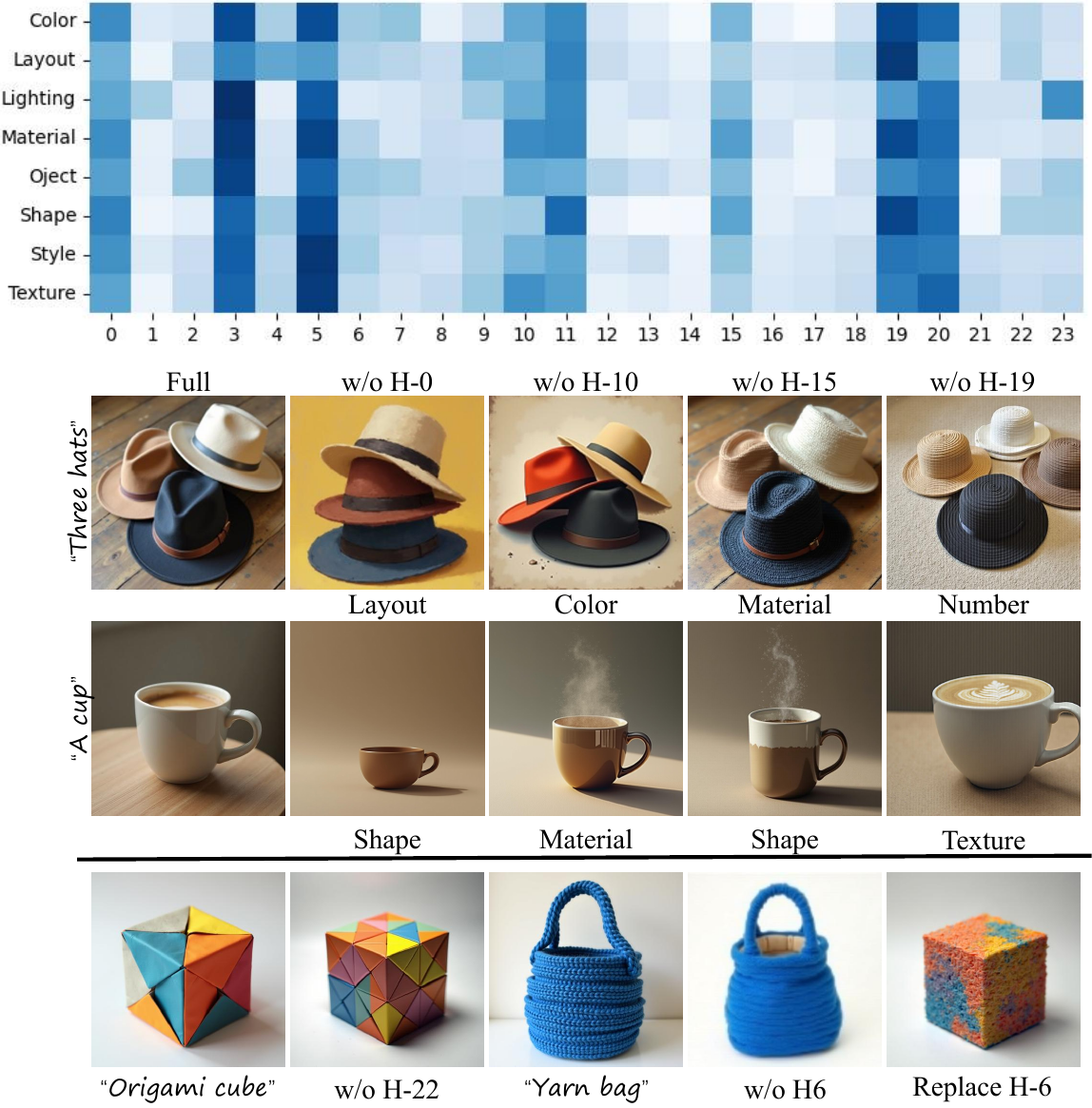}
  \caption{\textbf{Analysis of multi-head attention in MM-DiTs.} We illustrate the distribution of distinct semantics across attention heads. Dropping the most influential head leads to significant shifts in associated semantics while swapping the output features of this head enables targeted semantic injection to a certain degree.}
 \vspace{-3mm}
  \label{fig:analysis_head}
\end{figure}

\subsection{Multi-head attention semantic sensitivity}
Attention mechanisms have become fundamental components in Transformer architectures \cite{vaswani2017attention, dosovitskiy2020image}. Particularly, The multi-head attention mechanism enhances the representation power by allowing multiple attention heads to operate on different input projections. Formally,
\begin{equation}
    \text{MultiHead}(Q,K,V) = Concat(head_{1}, ...,head_{H})W^{o},
\end{equation}
where
\begin{equation}
    head_{h} = \text{Attention}(QW_{h}^{Q},KW_{h}^{K},VW_{h}^{V}).
\end{equation}
The $W^{o}$ is the projection for attention output, and $W_{h}^{Q}$, $W_{h}^{K}$, $W_{h}^{V}$ indicate projections for Query, Key and Value where $h$ indicates the head index and $H$ represents the total number of heads.
Previous studies have explored the role of multiple attention heads in models like the CLIP encoder \cite{radford2021learning}, revealing that different heads capture various correlations between text and images. This suggests that each head may specialize in representing specific semantic concepts or attributes within the multimodal data.

In this work, we  conduct an in-depth investigation into the influence of each individual attention head with regard to the representation of diverse editing semantics (e.g., shape, color, texture, style, etc.) in MM-DiTs. Specifically, We attempt to investigate whether certain attention heads are more sensitive to specific attributes and how this knowledge can be exploited for more effective image editing.

To this end, we construct a diverse paired-text dataset $\mathcal{D}$ to represent eight different editing semantics in the image editing benchmark (e.g., PIE-Bench~\cite{ju2024pnp}, TEDBench++~\cite{brack2024ledits++} and EditEval~\cite{huang2024diffusion}, etc.).
Let $S = \{ s_1, s_2, \ldots, s_{8} \}$ denote semantic categories, for each semantic category $s \in S$, we define its corresponding vocabulary set as $W_s$. For the vocabulary set excluding the current semantic category, we define $W_{-s} = W \setminus W_s$, where $W$ is the set of all vocabulary words. We then define the prompt construction function as:
\begin{equation}
      f: W_s \times W_{-s} \rightarrow P,
\end{equation}
which maps two words to a prompt. For example,  $f(w, u) = [\text{``a }, w, u\text{''}$]. Then, the corresponding subset of the dataset $D_s$ is defined as:
\begin{align}
    D_s =  (p_1, p_2) \mid p_1 = f(w_1, u_1),\ p_2 = f(w_2, u_2), \\
    \ w_1, w_2 \in W_s,\ w_1 \neq w_2,\ u_1, u_2 \in W_{-s}.
\end{align}
Here, $p_1$ and $p_2$ are prompts containing different words from the same semantic category $s$ and words from other semantic categories.
Finally, we combine all subsets to form the complete dataset $\mathcal{D}$:
\begin{equation}    
\mathcal{D} = \bigcup_{s \in S} D_s.
\label{eq:dataset}
\end{equation}
Each $D_s$ contains 500 pairs, so the entire dataset $\mathcal{D}$ contains 4000 pairs.
% Specifically, for each attribute (e.g., color), we create multiple text prompts that, in addition to the target attribute, also included other diverse attributes (e.g., material, shape, etc). These prompts are then used to guide image generation. By calculating the cosine similarity between corresponding heads during generation, we measure the common representation of attributes across different heads, thus assessing sensitivity of each head to specific attributes.
% To capture the output space of each head, we require a sufficiently diverse dataset. Therefore, we used GPT-4 to generate initial attribute pairs (e.g., material of color) and, based on these, manually prompt GPT-4 to generate more specific attribute variations (e.g., ceramics of blue), resulting in a total of 2,500 paired texts. 
We normalize each head’s sensitivity to different semantics and display the results as a heat map at the top of Fig.~\ref{fig:analysis_head}. The heat map discloses diverse sensitivities among heads, thereby inspiring us to adaptively activate particular heads, especially those with the highest sensitivities, so as to augment semantic representation.

Based on the above observation, we perform dropout on the heads that show the highest sensitivity for each semantic during inference and present results in the middle of Fig.~\ref{fig:analysis_head}. The findings indicate that dropout causes a significant change in the representation of the corresponding semantics, confirming the importance of these heads for specific semantics. Furthermore, we swap the output features of the most sensitive heads from two images. The results at the bottom of Fig.~\ref{fig:analysis_head} illustrate that through the exchange of these susceptible heads, specific semantics can be injected to a certain degree, which offers valuable insights into training-free semantic representation and editing.
%-------------------------------------------------------------------------------------------------

\subsection{Text-image token interactions}
\begin{figure}[!t]
  \centering
  \includegraphics[width=0.95\linewidth]{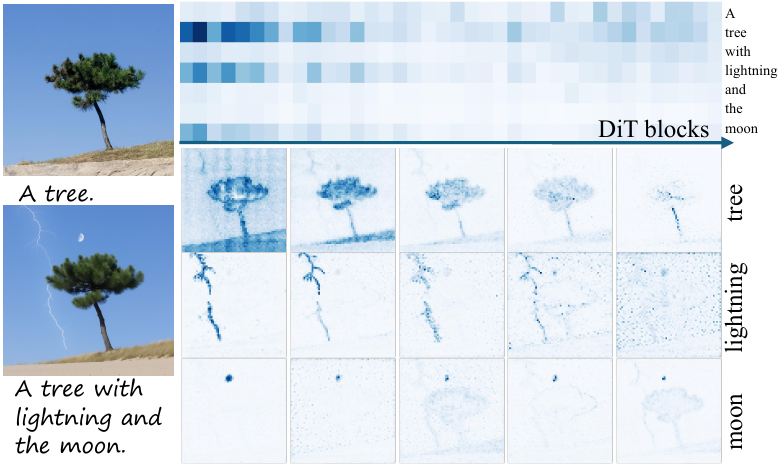}
  \vspace{-2mm}
  \caption{\textbf{Analysis of text guidance on image tokens.}  Key image regions influenced by text guidance are identified within the joint self-attention map and visualized. Additionally, we observe that text guidance influence diminishes as attention blocks progress in the denoising steps, leading to weakened semantic representation from text-guided editing.}
  \vspace{-4mm}
  \label{fig:analysis_token2}
\end{figure}
% In text-to-image diffusion models, guiding image editing through textual descriptions is a pivotal aspect.
Previous text-guided image editing methods mainly utilize cross-attention mechanisms to incorporate textual guidance~\cite{hertzprompt}.
Formally, in cross-attention, the features from the noisy image $\phi(z_{t})$ are projected to query $Q_{ca} = P_{Q}(\phi(z_{t}))$ and the text embedding $\psi(p)$ is projected to key $K_{ca} = P_{K}(\psi(p))$ and value $V_{ca} = P_{V}(\psi(p))$, where $P_{Q}$, $P_{K}$ and $P_{V}$ are pre-trained linear projections. 
Then, an attention map can be explicitly obtained via:
\begin{align}
    \text{Attention map}(Q_{ca}, K_{ca}, V_{ca}) = \text{softmax}\left(\frac{Q_{ca}K_{ca}^T}{\sqrt{d_k}}\right),
\end{align}
and can be directly used for various operations.

However, MM-DiT presents a distinct paradigm where text and image token embeddings are combined into a single input sequence and then processed by transformer blocks using joint attention. Formally, the input to the joint attention can be formalized as follows:
\begin{align}
    % \text{Joint embeddings}(z_{t}, p_{t}) = \phi(z_{t}) \odot \psi(p_{t}),
    Q = P_{Q}^{I}(\phi(z_{t})) \odot P_{Q}^{T}(\psi(p_{t})), \\
    K = P_{K}^{I}(\phi(z_{t})) \odot P_{K}^{T}(\psi(p_{t})), \\
    V = P_{V}^{I}(\phi(z_{t})) \odot P_{V}^{T}(\psi(p_{t})),
\end{align}
% $z_{t}$ and $p_{t}$ are the input of noised image and text prompt, respectively. $\phi(z_{t})$ and $\psi(p_{t})$ are the feature embeddings of image and text, 
where $P_{Q}^{I}$, $P_{K}^{I}$, $P_{V}^{I}$, $P_{Q}^{T}$, $P_{K}^{T}$ and $P_{V}^{T}$ are pre-trained linear projections for image and text embeddings and $\odot$ indicates image embedding and text embeddings are concatenated in the token length dimension. Joint attentions in MM-DiT are as follow:
\begin{align}
    \text{Attention}(Q, K, V) = \text{softmax}\left(\frac{QK^T}{\sqrt{d_k}}\right)V.
\end{align}
This joint attention among text and image tokens allows for more intricate interactions and makes influence dynamics less explicit.

% We focus on the influence of text tokens on image tokens within the joint framework of MM-DiT. 
% We first explore how text tokens guide image tokens during generation. Specifically, attention weights are calculated via a scaled dot-product followed by a cross-modal $softmax$ (calculate pre-row in joint self-attention map); we then locate the down-left part of the joint self-attention maps where specific text tokens of each column most strongly influence image tokens of each row. We further extract these attention weights and reshape them into a 64×64 grid to create a heatmap. Results presented in Fig.~\ref{fig:analysis_token2} demonstrate that image tokens are attentive to the words that describe them in this region. The heatmap shows a strong correlation between text and image tokens, where the semantic textual descriptions have a stronger impact on the corresponding image regions.

% We then explore the variations of text guidance in MM-DiT, highlighting a key difference from UNet-based models. In UNet-based models, text is encoded initially, providing fixed guidance for image generation across all cross-attention layers. In contrast, MM-DiT incorporates text embeddings via joint self-attention, updating them progressively across attention blocks. We quantify the attention weights of each text token across attention blocks, as shown in Fig.~\ref{fig:analysis_token2}, and observe that text guidance decays with deeper blocks, gradually weakening its effect on image generation.

Our study delves into the influence of text tokens on image tokens within the framework of MM-DiT. 
Firstly, we conduct an analysis of the manner in which text tokens direct influence image tokens during the generation process. 
This is achieved by computing attention weights through the utilization of a scaled dot-product along with a cross-modal $softmax$ across the rows within the joint self-attention map. 
Our focus is centered on the lower-left area of this map, where each column reveals the impact of every word, and each row demonstrates how the intent of each image token is influenced.
Subsequently, we reshape the weights into a 64×64 grid for the purpose of constructing a heatmap.
As shown in Fig.~\ref{fig:analysis_token2}, The heatmap discloses that image tokens are responsive to relevant textual descriptions, thereby signifying a robust alignment between text and image tokens within this particular region.
%We first analyze how text tokens guide image tokens during generation by calculating attention weights using a scaled dot-product and a cross-modal $softmax$ across rows in the joint self-attention map. We focus on the lower-left region of this map, where each text token in a column strongly influences each image token in a row. We then reshape the weights into a 64×64 grid to form a heatmap. As shown in Fig.~\ref{fig:pipeline}, the heatmap reveals that image tokens are attentive to relevant textual descriptions, indicating a strong alignment between text and image tokens in this region.

Next, we examine the dynamics of text guidance in MM-DiT, emphasizing a key contrast with UNet-based models. 
%In UNet-based models, text is encoded only once, providing fixed guidance through all cross-attention layers. In MM-DiT, however, text embeddings are integrated through joint self-attention, hence changing progressive along with attention blocks. 
In UNet-based models, the text is encoded merely once, thereby furnishing fixed guidance throughout all cross-attention layers. Conversely, in MM-DiT, the text embeddings are incorporated via joint self-attention, and thus they evolve progressively in tandem with the attention blocks.
By measuring each text token's attention weight across blocks, we obverse that the text guidance wanes with increasing blocks depth, gradually lessening its impact on the process of image generation, Fig.~\ref{fig:analysis_token2}.

% 总结
% This analysis underscores the effectiveness of the self-attention mechanism in aligning textual and visual information within multimodal diffusion transformers. By understanding how text tokens influence image tokens, we can enhance text-guided image editing methods to achieve more precise and semantically consistent results. The entangled representations facilitate a richer interaction between modalities, enabling the generation of images that are more faithfully guided by textual descriptions. 
\section{Method}
\label{sec:method}
\begin{figure*}[!h]
  \centering
  \includegraphics[width=0.95\linewidth]{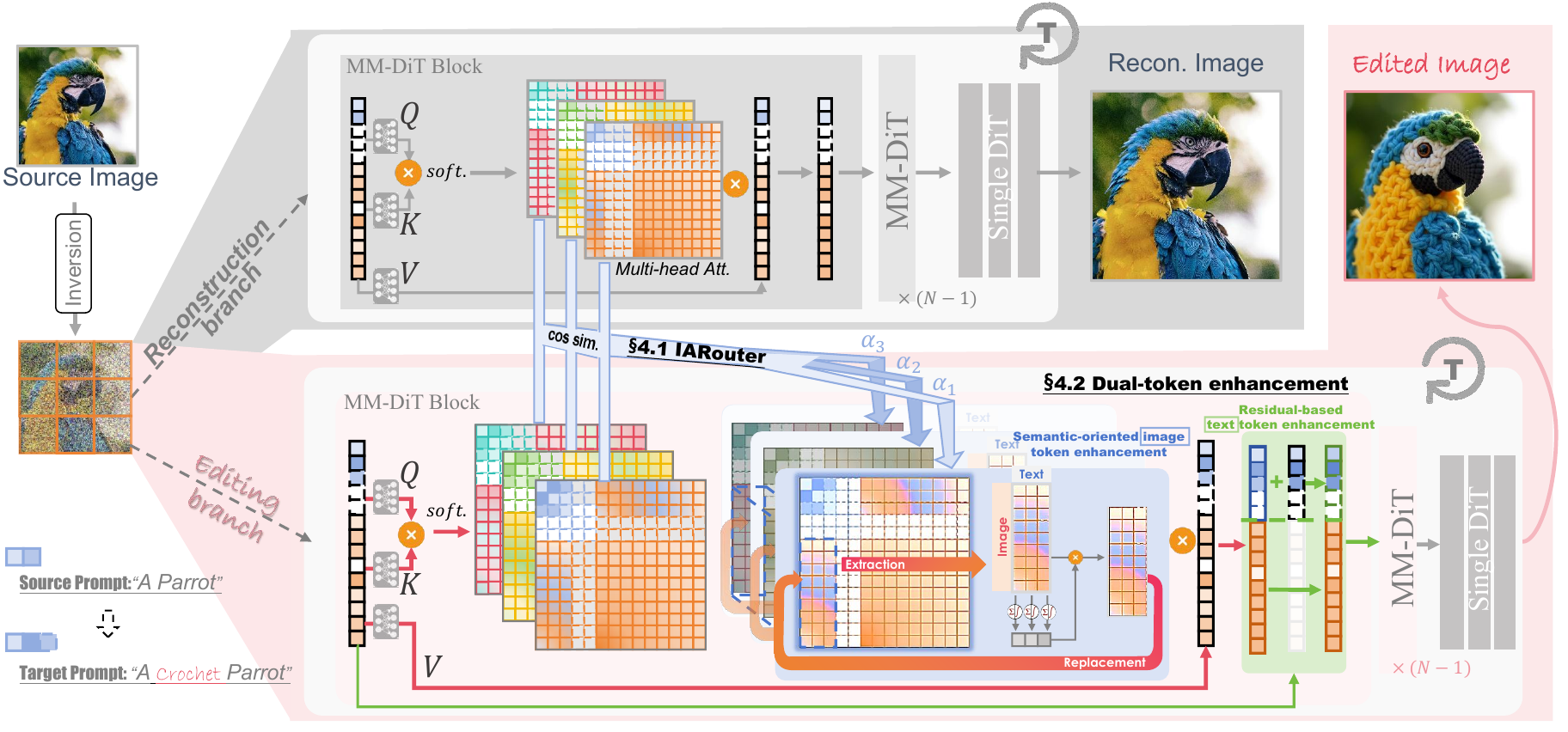}
  \caption{\textbf{Pipeline of our method.} We mainly introduce instance-adaptive attention head router \textbf{(IARouter)} to adaptively activate attention heads based on their semantic sensitivity, enabling a more accurate representation of the edited specific images.}
  \label{fig:pipeline}
\end{figure*}
% In this section, we present our methods to enhance text-guided image editing based on our analysis  of attention mechanisms in MM-DiT. We mainly propose two techniques:
% \textbf{Attribute-focused head enhancement} which enhancing the representation of desired editing attributes by identifying and emphasizing the most effective attention heads, and \textbf{Token-Aware enhancement} which focusing edits on key image tokens by utilizing attention weights between text and image tokens.

We primarily propose two techniques: \textbf{Instance-adaptive Attention Head Router}, which improves the representation of target editing semantics by identifying and emphasizing the most effective attention heads; and \textbf{Dual-token refinement module (DTR)}, which refines edits on key image tokens by applying attention weights from text to image tokens.

\subsection{Instance-adaptive Attention Head Router}
% \subsection{Adaptive attention head weighting}
% In this section, we introduce our method to enhance the representation of desired editing attributes in diffusion Transformers. 
Building upon our analysis of attention heads’ sensitivities to different editing semantics, we aim to identify and emphasize the most effective attention heads for specific editing tasks. By leveraging information from the image reconstruction branch, we guide the image editing branch to focus on the most relevant attention heads, thereby improving editing effectiveness.

We first identify effective attention heads.
The key to our approach is identifying which attention heads are most sensitive to the desired editing semantics. Given a DiT model with $H$ attention heads, we begin by calculating the cosine similarity between the outputs of corresponding attention heads when generating images with and without a specific semantic. Let $\mathbf{v}_{h}^{(r)}$ and $\mathbf{v}_{h}^{(e)}$ denote the output features of the $h$-th attention head in the reconstruction branch and the editing branch, respectively, the cosine similarity $s_{h}$ for head $h$ is calculated as:
\begin{equation}
s_h = \frac{\mathbf{v}_h^{(r)} \cdot \mathbf{v}_h^{(e)}}{|\mathbf{v}_h^{(r)}| |\mathbf{v}_h^{(e)}|},
\end{equation}
where  $\cdot$  denotes the dot product and  $\| \cdot \|$  represents the Euclidean norm.

To quantify the sensitivity of each attention head to the specific semantic and normalize the dissimilarity scores in a single step, we design the normalized dissimilarity score $\tilde{d}_h$ for head $h$ as:
\begin{equation}
\tilde{d}h = \frac{s_{\text{max}} - s_h}{s_{\text{max}} - s_{\text{min}}},
\label{eq:dissimilarity}
\end{equation}
where,
\begin{equation}
s_{\text{min}} = \min\{s_1, s_2, \dots, s_H\},
\end{equation}
\begin{equation}
s_{\text{max}} = \max\{s_1, s_2, \dots, s_H\}.
\end{equation}

% A lower cosine similarity indicates a greater difference between the attention head outputs, suggesting that the head is more sensitive to the editing attributes. Therefore, attention heads with lower $s_{i}$ values are considered more effective for the desired edits.
This normalized score $\tilde{d}_h$ reflects how dissimilar each attention head’s outputs are with respect to the specific semantics, relative to the range of dissimilarities observed across all heads.

To smoothly activate the most semantic-sensitive attention heads, we propose an instance-adaptive attention head router (\textbf{IARouter}) applied to the output features of different heads. IARouter is designed to (1) highlight dissimilar heads: assign high attention to heads with lower $\tilde{d}_h$ to emphasize their importance in representing the desired editing semantics; (2) maintain similar heads: ensure that the contributions of heads that are less relevant to the edits are not excessively altered, thereby maintaining the integrity of other visual aspects in the image; (3) smooth weights: avoid artifacts by preventing sudden weight changes and maintain model stability.
	
% 2.	Promote sparsity: Ensure that the weighting is sparse, meaning that only the most dissimilar heads are significantly weighted, while the features from other heads remain largely unchanged.

Based on these objectives, IARouter uses soft activate attention heads. The weight $w_h$ for head $h$ is defined as:

\begin{equation}
w_h = 1 + \gamma \cdot \sigma\left( k (\tilde{d}_h - \delta) \right),
\label{eq:head}
\end{equation}
% gamma=0.5, k=10, delta=0.5
where $\gamma$ is the maximum weight increment, $k$ controls the steepness of the sigmoid curve, $\delta$ shifts the center of the sigmoid, and $\sigma(x)$ is the sigmoid function defined as:

\begin{equation}
\sigma(x) = \frac{1}{1 + e^{-x}}.
\end{equation}

% By adjusting $\gamma$, $k$, and $\delta$, we can fine-tune the weight distribution to achieve the desired level of amplification for the most dissimilar heads.

During the image generation process, we multiply the output of each attention head by its corresponding weight to obtain the enhanced output:
\begin{equation}
\mathbf{v}_h^{\text{enhanced}} = w_h \cdot \mathbf{v}_h^{\text{e}}.
\end{equation}

% This weighting process enhances the contributions of the most effective attention heads while keeping the influence of other heads largely unchanged, improving the model’s ability to perform the desired edits.

% Theoretical Justification

% The proposed AFHE weighting scheme aligns with the principles of attention mechanisms by modulating the influence of different components based on their relevance to the task.
% By focusing on the attention heads that are most responsive to the editing instructions, we enhance the model’s ability to incorporate the desired changes while preserving unaffected attributes.

% The sparsity in the weight distribution ensures that we do not excessively alter the contributions of heads that are less relevant to the edits, thereby maintaining the integrity of other visual aspects in the image.
The proposed IARouter serves as a smooth semantic-specific enhancer.
By identifying and emphasizing heads sensitive to specific semantics, IARouter allows for more precise and effective edits.
Using a sigmoid function allows for a gradual increase in weights, preventing sudden changes that could introduce artifacts.
% (2) Flexibility: The approach can be adapted to various editing tasks by adjusting the hyperparameters and can be extended to other transformer-based models.

%可以用在intro或者conclusion中
% Conclusion
% Our proposed method enhances text-guided image editing by leveraging the attribute sensitivities of individual attention heads within the diffusion transformer. By adaptively weighting the heads most responsive to the desired edits, we achieve more effective and targeted image manipulation. This approach contributes to the advancement of fine-grained control in text-to-image diffusion models, enabling the generation of images that more faithfully reflect the specified editing attributes.

% 双端的token增强方法；important text token leads to imaportant image token
\subsection{Dual-token Refinement Module}
\label{subsec:duab-token}
Sec.~\ref{sec:analysis} shows that attention weights between text and image tokens reflect the influence of the text prompt on each image token. We utilize these weights to focus edits on key image regions corresponding to the desired semantics for semantic refinement. Additionally, we propose modifying the attention normalization to amplify the impact of significant text tokens on the image tokens.

% Leveraging Attention Weights for Targeted Editing
% \paragraph{Focusing Edits on Key Image Tokens}
\paragraph{Semantic-oriented image token enhancement.}The self-attention mechanism in MM-DiTs produces attention weights that reflect the influence of text tokens on image tokens. Specifically, for each image token, the attention weights associated with text tokens indicate the degree to which that image token attends to each text token. We exploit this property to identify and focus on image tokens that are most affected by the editing prompt.
Let  $\mathbf{A} \in \mathbb{R}^{N \times M}$  denote the attention weight matrix from text tokens to image tokens, where  $N$  is the number of image tokens and  $M$  is the number of text tokens. The element $\mathbf{A}_{i,j}$ represents the attention weight from text token $j$ to image token $i$. 
% We extract the $j$-th column in $\mathbf{A}_{ij}$ and get attention weights, 
% we compute a weight  $w_i$  for each image token by aggregating its attention weights over all text tokens:
% \begin{equation}
% w_i = \sum_{j=1}^{M} \mathbf{A}_{ij}
% \end{equation}
% We normalize these weights to ensure they lie within the range [0, 1]:

% \begin{equation}
% \hat{w}_i = \frac{w_i - \min(w)}{\max(w) - \min(w)}
% \end{equation}
% where  $\min(w)$ and  $\max(w)$  are the minimum and maximum values in the set $\{ w_i \}$.

We propose \textbf{semantic-oriented image token enhancement} to focus the editing on key image tokens, taking into account the impact of text on different image tokens. Formally, the weight mapping is defined as:
\begin{equation}
      \hat{w}_{i,j} = \alpha \cdot \sigma \left(\upsilon \cdot  \frac{e^{\mathbf{A}_{i,j}}}{\sum_{k=1}^{N} e^{\mathbf{A}_{k,j}}} \right),
      % \hat{w}_{i,j} = \ln \left(1 + \upsilon \cdot  \frac{e^{\mathbf{A}_{i,j}}}{\sum_{k=1}^{N} e^{\mathbf{A}_{k,j}}} \right),
    \label{eq:tar}
\end{equation}
where $\mathbf{A}_{i,j}$ indicates the attention weight of $j$-th text token to $i$-th image token.
We using a $softmax$ based function that normalizes attention weights of image tokens, and using sigmoid functions to limit the growth of large weights. $\alpha$ is a weight enhancement coefficient, and $\upsilon$ is for amplitude adjustment. Further discussion on influence of $\alpha$ and $\upsilon$ can be found in the supplementary materials.

Next, we reweight the image tokens in the editing branch using the normalized weights  $\hat{w}_{i,j}$. The final image tokens  $\mathbf{\hat{A}}_{i,j}$ are computed as:
\begin{equation}
\mathbf{\hat{A}}_{i,j} = \hat{w}_{i,j} \cdot \mathbf{A}_{i,j}.
\end{equation}
This formulation ensures image tokens highly influenced by the text prompt (with higher $\hat{w}_{i,j}$) are assigned high weights, while tokens less influenced remain close to the original.

\paragraph{Residual-based text token enhancement.}
As attention weights between text and image tokens decay through successive attention blocks, we leverage residual text tokens to retain text guidance in each transformer block. Specifically, this design carries over text guidance from the previous attention block to the current one. As information progresses through deeper blocks, the previous block’s input is incorporated as the residual term and combined with the current block’s input, enhancing the continuity of text guidance. This mechanism introduces consistent text information into each block, strengthening text guidance and improving the accuracy of image editing.
\section{Experiments}
\label{sec:experiments}
\begin{figure}[!t]
  \centering
  \includegraphics[width=0.97\linewidth]{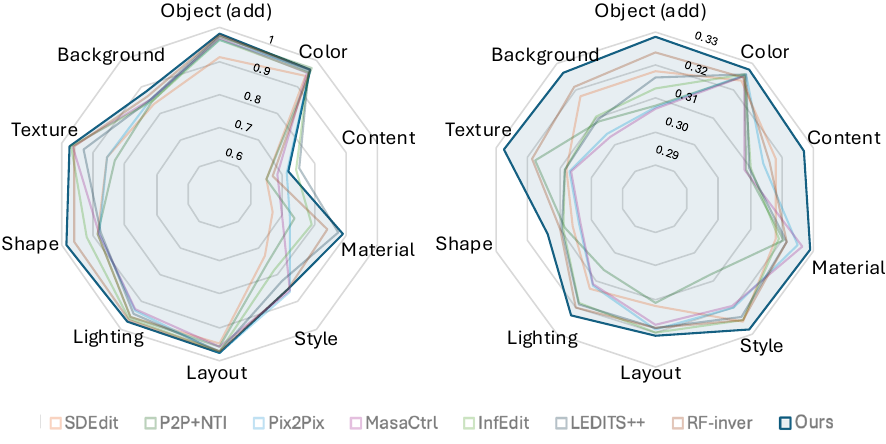}
  \vspace{-3mm}
  \caption{\textbf{Radar chart for evaluating image and prompt alignments in eight editing tasks.} Overall, our approach effectively retains the intrinsic feature of the original image while aligning precisely with the specified text guidance. }
  \label{fig:chart}
    \vspace{-3mm}
\end{figure}
% surpasses the current SOTA methods across diverse image editing tasks,
\begin{table*}[!th]
\centering

\resizebox{\linewidth}{!}{%
\begin{tabular}{c|ccccccc|c}
\toprule
Methods   &SDEdit &P2P+NTI     &Pix2Pix &MasaCrtl &InfEdit &LEDITS++   &RF-Inver.     &Ours     \\ \hline
Structure-alignment (DINO) $(\uparrow)$   	&0.8377 &0.8559	&0.8722	&0.8744	&0.8909	&0.8963	&0.9029 &\textbf{0.9194}\\ \hline
Prompt-alignment (CLIP) $(\uparrow)$  	&0.3019	&0.2944	&0.2975	&0.2955	&0.3016	&0.3022	&0.3098 &\textbf{0.3203}\\  \hline
Image-quality (LPIPS) $(\downarrow)$  &0.2151 &0.2258 &0.2975 &0.3144 &0.2702 &0.2796 &0.2262 &\textbf{0.2103} \\ \bottomrule
\end{tabular}
}
\caption{\textbf{Comparison of cosine similarity between DINO features (for image) and CLIP (for text) of the edited images and source images and prompts, respectively}. Our method has the best scores, indicating that our approach successfully edit the image guided by text while maintain consistency with source image and high image quality.}
\label{tab:alignment}
\end{table*}
\begin{figure*}[!h]
  \centering
  \includegraphics[width=0.97\linewidth]{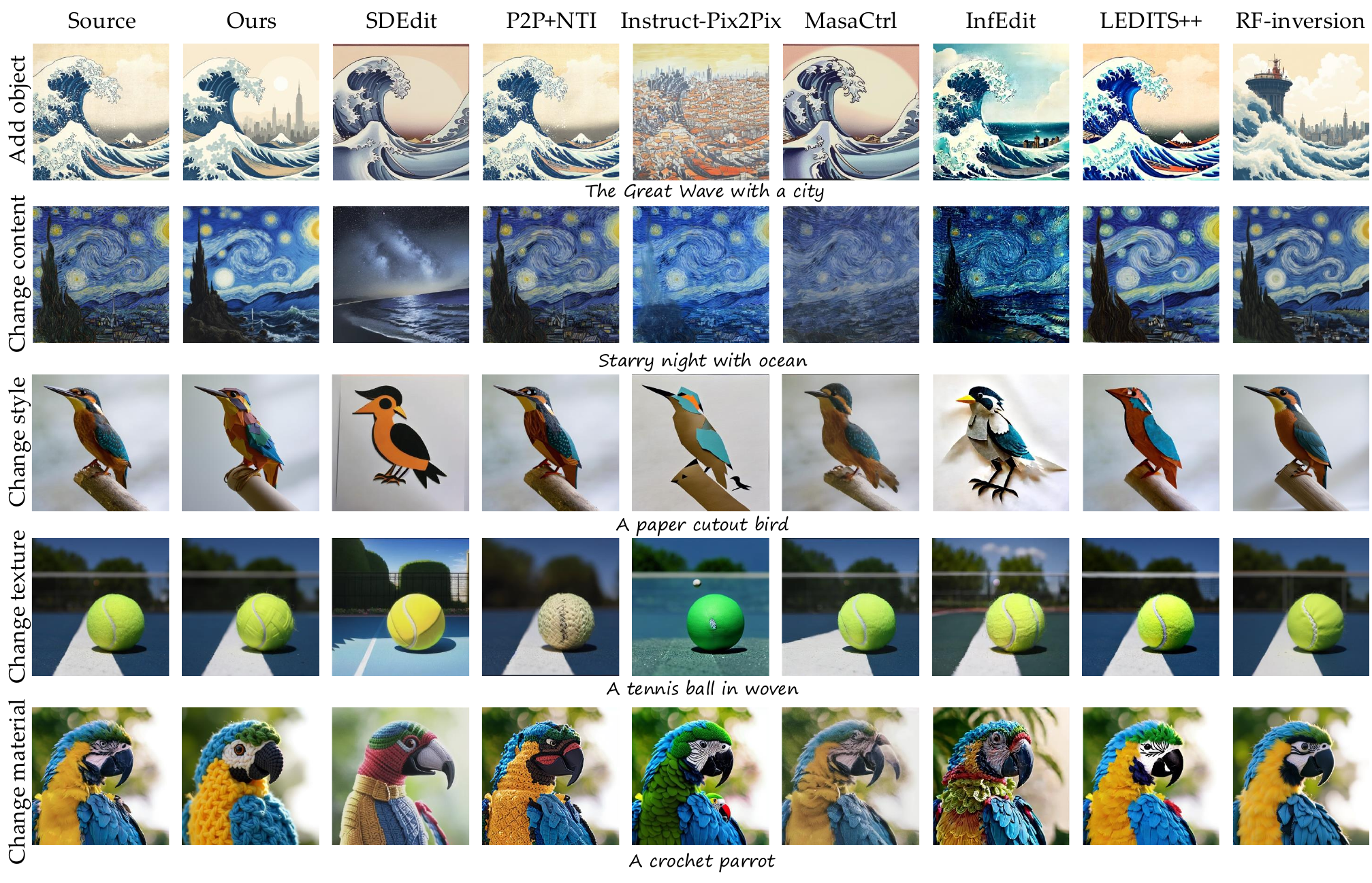}
 \vspace{-3mm}
  \caption{\textbf{Qualitative comparison with baseline methods on various editing tasks.} Our results demonstrate high alignment with the text guidance while keeping consistency with the reference image.}
   \vspace{-3mm}
  \label{fig:fig_main_compare}
\end{figure*}
\paragraph{Implementation details.}
In our experiments, we utilize Flux-1.0[dev]~\cite{flux} with default hyperparameters, we leverage RF-Inversion~\cite{rout2024semantic} to inverse a real image to its latent space, and basic settings are followed with them.

\paragraph{Baselines.}
We compare our method with seven state-of-the-art text guided image editing approaches, include two Flux based approaches: RF-inversion~\cite{rout2024semantic} and SDEdit~\cite{meng2021sdedit} and five UNet based approaches: Null-textual Inversion~\cite{mokady2023null} with prompt-to-prompt~\cite{mokady2023null}, Instruct-Pix2Pix~\cite{brooks2023instructpix2pix}, MasaCtrl~\cite{cao2023masactrl}, InfEdit~\cite{xu2024inversion}, and LEDITS++~\cite{brack2024ledits++}, all the approaches are training free.

\paragraph{Datasets.}
We evaluate our method with baselines on two text-guided image editing benchmarks TEDBench++~\cite{brack2024ledits++}, which is the revised extension of TEDBench~\cite{kawar2023imagic} and contains 120 entities in total, PIE-Bench~\cite{ju2024pnp}, which comprises of 700 images, each associated with ten distinct editing types.

\paragraph{Evaluation Metrics.}
Following previous text-guided image editing work~\cite{nam2024contrastive, brack2024ledits++, xu2024inversion, huberman2024edit}, we evaluate the proposed method across three metrics: results of overall image quality, alignment with text guidance, and structure consistency with source images. Specifically, we use LPIPS~\cite{zhang2018unreasonable} to assess overall quality, CLIP-T~\cite{ilharco_gabriel_2021_5143773} to measure text alignment, and DINO~\cite{oquab2023dinov2} to evaluate structure consistency with the original image. Additionally, we conduct a user study to further assess performance.

\subsection{Qualitative comparison}
In Fig.~\ref{fig:fig_main_compare}, we present the visual results of different editing types in comparison with baseline methods. 
SDEdit~\cite{meng2021sdedit} is capable of generating new concepts within textual conditions (e.g., ocean, paper cutout), but it struggles to maintain the semantic information of source images (in the $2^{nd} and 3^{rd}$ rows).
% , or it is challenging to integrate new concepts into the reference images (in the $1^{st}$ row).
P2P+NTI~\cite{mokady2023null} is difficult to achieve satisfactory image editing results, often neglecting the information contained within textual conditions (in the $1^{st} \sim 3^{rd}$ rows).
Instruct-Pix2Pix~\cite{brooks2023instructpix2pix} also struggles with image editing instructions that involve significant changes, leading to semantic loss (in the $1^{st}$ row) or inaccurate editing (in the $2^{nd}$, $4^{th}$, and $5^{th}$ rows).
MasaCtrl~\cite{cao2023masactrl} and InfEdit~\cite{xu2024inversion} similarly fail to accurately preserve the semantics of source images (in the $3^{rd}$ row) and inaccurate editing (in the $1^{st}$, $2^{nd}$, and $5^{th}$ rows).
LEDITS++~\cite{brack2024ledits++} achieves editing effects in experiments that alter the style of images. However, there is still an issue with specific semantic editing (in the $1^{st}$, $2^{nd}$, $4^{th}$, $5^{th}$rows) the loss of details in source images (in the $3^{rd}$ row).
RF-inversion~\cite{rout2024semantic} struggles to achieve robust image editing effects, resulting in some outputs that are mostly the same as the Source images (in the $3^{rd}$ and $5^{th}$ rows).
% In comparison, our method is capable of integrating concepts reasonably, replacing the original objects in the source image.
Our approach achieves the best structural preservation and editing effects, surpassing the performance of baseline methods.

\subsection{Quantitative comparison}
Tab.~\ref{tab:alignment} presents the quantitative comparison, including image structure alignment between the edited image and the source image, alignment between the edited image and text guidance, and overall generation quality.
We further evaluate image and text alignment across eight different editing types, as shown in the radar chart in Fig.~\ref{fig:chart}. 
In the ``change content'' category, although InfEdit and LEDITS++ have comparable metrics to ours, their text alignment is significantly lower, indicating that these methods fail to achieve effective ``change object'' editing. Similar conclusions can be observed in Fig.~\ref{fig:fig_main_compare}. Besides, the ``change content'' score is significantly lower than other metrics because content changes introduce substantial alterations in the main areas of the images, reducing structure similarity with the original images. Nonetheless, our results are still considerably better when compared to baseline outcomes.

\begin{figure}[!t]
  \centering
  \includegraphics[width=0.8\linewidth]{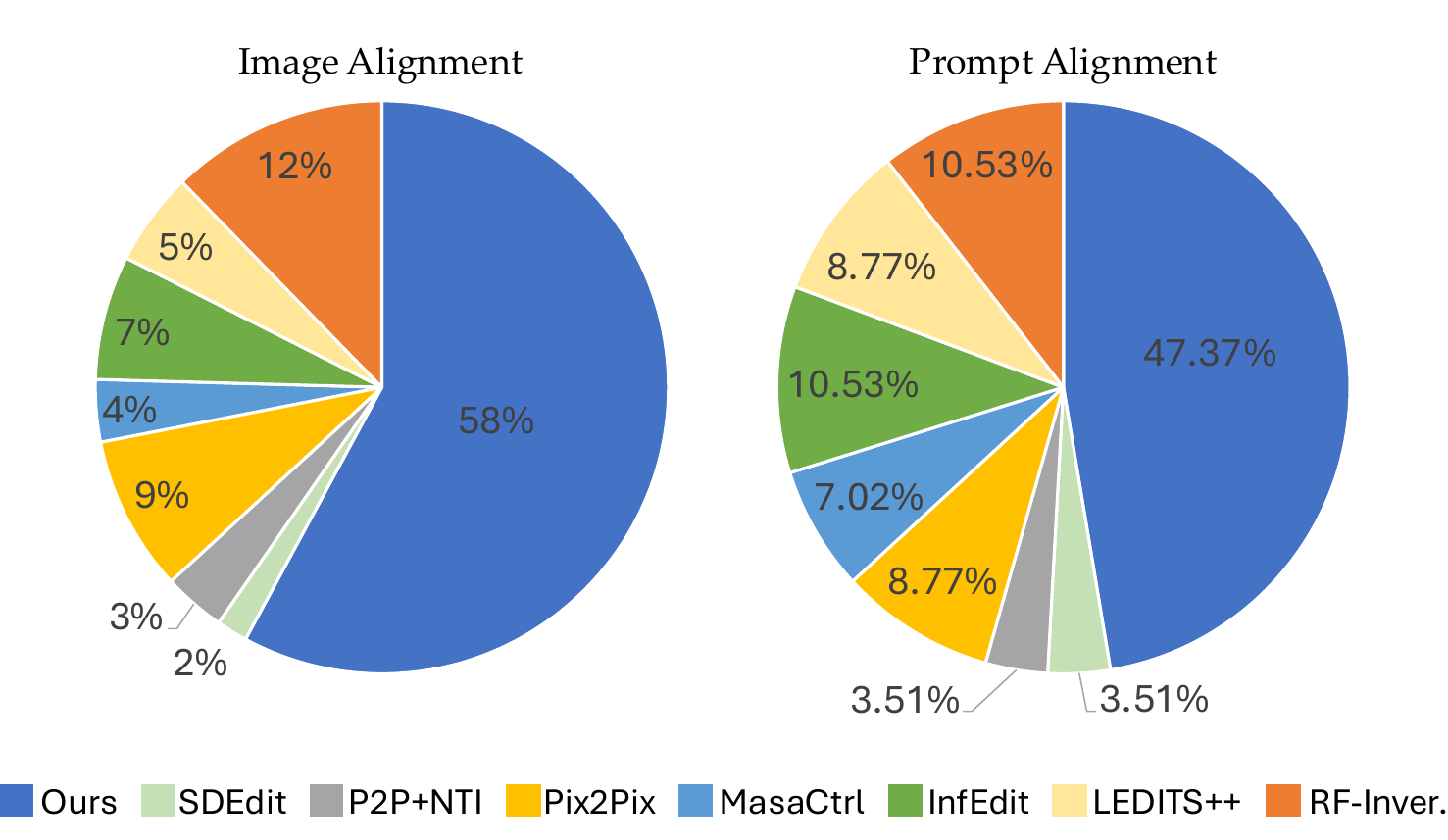}
  \vspace{-2mm}
  \caption{\textbf{Statistics for user study on alignments.}}
  \vspace{-4mm}
  \label{fig:user_study}
\end{figure}

\paragraph{User study.}
We conduct a user study focusing on two primary aspects: alignment with the given prompt and preservation of irrelevant regions in the image. We generated 50 groups of images across various editing tasks, each containing eight images generated by our method and seven generated by baseline methods using the same prompt.
56 participants were presented with each group of images and asked to select the image that best aligns with the prompt and the original image. Results of Fig.~\ref{fig:user_study} indicate that the results of our method closely follow the prompt while retaining the quality of regions not relevant to the editing prompt.
\subsection{Ablation study}
\begin{figure}[!t]
  \centering
  \includegraphics[width=0.97\linewidth]{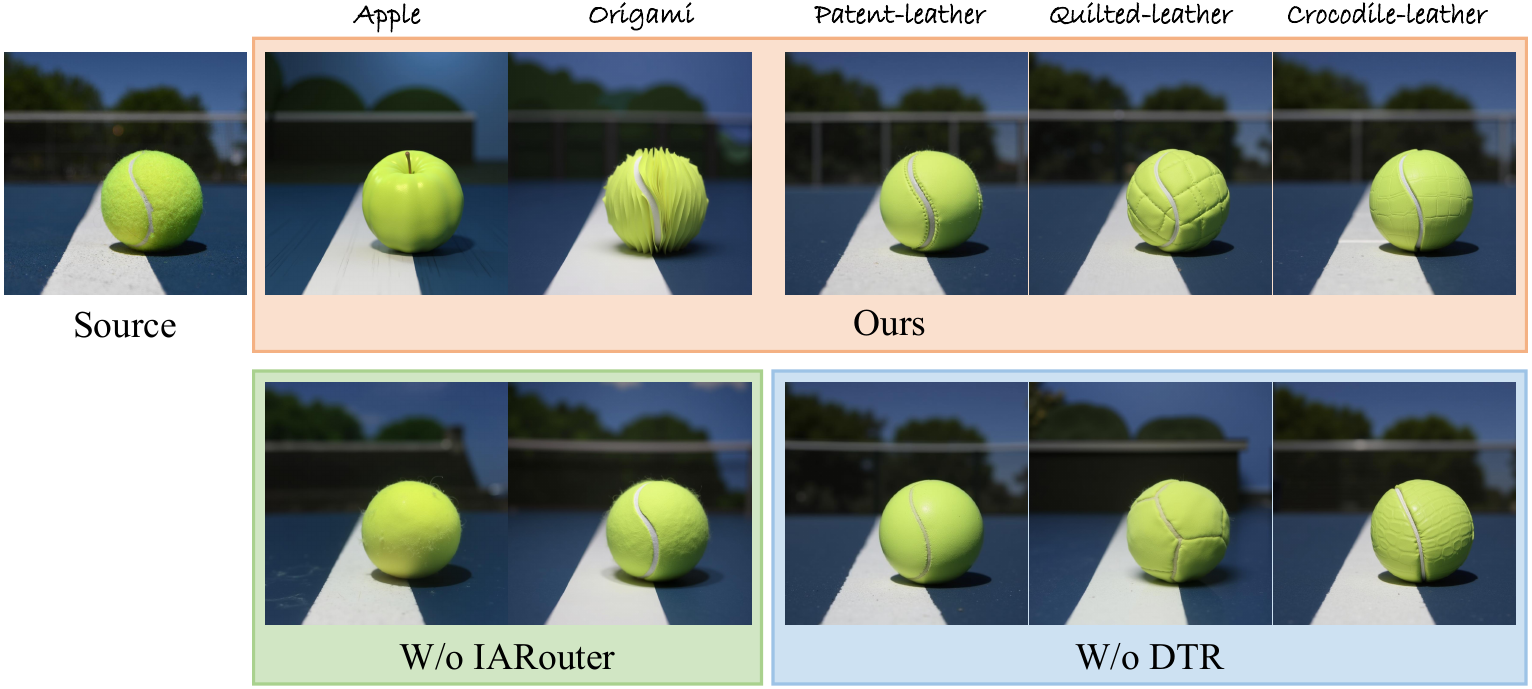}
  \vspace{-2mm}
  \caption{\textbf{Visual analysis for IARouter and DTR.} Without \textbf{IARouter}, results tend to weaker semantic representation and retain the original semantics. With \textbf{DTR}, we achieve more fine-grained semantic representation. }
  \vspace{-5mm}
  \label{fig:ablate}
\end{figure}

In this section, we validate the effectiveness of our approach by ablating the two key modules we propose: instance-adaptive attention head router \textbf{(IARouter)} and dual-token refinement module (\textbf{DTR}). First, for the IARouter ablation, we remove any head constraints during inference. As shown in the lower left of Fig.~\ref{fig:ablate}, this results in weaker semantic expression (e.g., in the ``apple'' and ``origami'' examples, although the desired editing semantics are somewhat achieved, residual textures from the original tennis ball image persist). In contrast, IARouter enhances the expression of specific semantics by routing different heads according to semantic content. Next, we ablate DTR, with results shown in the lower right of Fig.~\ref{fig:ablate}. The results indicate that by strengthening image tokens and text guidance, our method captures the desired semantics and achieves finer-grained semantic representation in response to detailed text guidance. 
\textbf{More ablation studies are provided in the supplementary material.}
%Without \textbf{DTR}, results tend to weaker semantic representation and retain the original semantics. With \textbf{DTR}, we achieve more fine-grained semantic representation.
\subsection{Limitation}
\label{sec:badcase}

% limitation:背景还是有改变，是因为inversion方法的问题
% Exchange specific heads can achieve feature inject.
% dynamic text embedding leads to feature weaken representation

Due to the multimodal text-image priors in the pre-trained models, when editing common elements like the ``Eiffel Tower'' with ``a $<$description$>$ Eiffel Tower''may yield limited results, as these prompts already encode specific visual details. Additionally, our approach requires image inversion to latent space, so the alignment between the edited result and the original image depends on the accuracy of the inversion process. \textbf{Further discussions on limitations and bad cases can be found in the supplementary materials.}

% Although our method achieves prompt-aligned editing of real images, it still encounters certain failure cases. Due to the limitations of multimodal text-image priors in the pre-trained models, suboptimal results may arise when editing samples with frequent and fixed description in the dataset. For instance, editing an image of the Eiffel Tower using ``a xxx Eiffel Tower"" may produce limited changes, as the prompt inherently encodes specific visual features (such as material and structure) of the Eiffel Tower from the pre-trained model. Additionally, our approach requires inversion of real images into the latent space for editing, meaning the accuracy of the inversion algorithm directly influence the alignment between the edited result and the original image. Detailed discussions of failure cases and limitations are provided in the supplementary materials.

\section{Conclusion}
\label{sec:conclusion}

% limitation:背景还是有改变，是因为inversion方法的问题
% Exchange specific heads can achieve feature inject.
% dynamic text embedding leads to feature weaken representation

In this work, we investigate multi-head attention within MM-DiTs for image editing, revealing the distribution of distinct image semantic information across heads. Additionally, we analyze text-to-image token guidance, observing that text influence diminishes in deeper attention blocks. Building on these insights, we introduce the instance-adaptive attention head router to enhance the representation of key attention heads for targeted editing semantics and propose the dual-token refinement module to ensure precise text guidance and emphasis on key regions. Extensive quantitative and qualitative evaluations, as well as user studies, demonstrate the superiority of our approach over existing state-of-the-art methods. 

\noindent \textbf{Furture work.} Future exploration could focus on leveraging images as guidance. Images inherently provide richer information, spanning from coarse-grained to fine-grained details. To achieve this, adapters for DiTs can be employed to effectively capture image-based information, enabling more consistent and semantically coherent image editing.
{
    \small
    \bibliographystyle{ieeenat_fullname}
    \bibliography{main}
}
\end{document}